# GLaM-Sign: Greek Language Multimodal Lip Reading with Integrated Sign Language Accessibility

DIMITRIS KOUREMENOS[1,2]
KLIMIS NTALIANIS[2]
[1]Institute of Informatics and Telecommunications,
NCSR "Demokritos",
Athens,
GREECE
[2]Department of Business Administration,
University of West Attica,
12243 Athens,
GREECE

*Abstract:* - The paper introduces the **Multimodal Greek Sign Language and Lip Reading Dataset v1.0 (GLaM-Sign)**[i] [1], developed under the Feelit project [2], aiming to enhance accessibility for Deaf and Hard-of-Hearing (DHH) individuals in the tourism sector. This multimodal dataset integrates synchronized audio recordings, video lip movements, text transcriptions, and Greek Sign Language (GSL) translations, supporting advanced applications in speech recognition, lip-reading AI, and real-time captioning. By addressing linguistic diversity and phonetic complexity, it sets a benchmark for accessibility-focused AI tools, fostering inclusivity and breaking communication barriers. The dataset aligns with cutting-edge innovations in multimodal AI while offering a culturally tailored solution for the Greek language. Despite challenges in word-level timestamp accuracy, the paper outlines future enhancements to optimize the dataset's usability for diverse research and applications. This resource represents a transformative step towards creating inclusive AI technologies for underrepresented communities.

## 1 Introduction

In recent years, advancements in digital learning tools have unlocked vast opportunities to enhance education across various disciplines, including vocational training. The FEELIT project emerges as a groundbreaking initiative that leverages cutting-edge technologies, such as Virtual Reality (VR) and multimodal datasets, to develop innovative resources aimed at improving the skills of professionals in the tourism industry and fostering inclusivity for Deaf and Hard-of-Hearing (DHH) individuals.

Tourism plays a pivotal role in the global economy, contributing approximately 10% to the global GDP and supporting millions of jobs worldwide. Despite its economic significance and its potential for cultural exchange, inclusivity challenges persist, particularly for DHH individuals. Barriers in communication, lack of accessible services, and limited training for tourism professionals often result in exclusion and inequity within the sector. Recognizing these issues, the FEELIT project seeks to bridge these gaps by equipping professionals with the tools and knowledge needed to provide accessible and inclusive services, fostering a more equitable tourism experience.

As part of its multifaceted approach to accessibility, the FEELIT project has developed the GLaM-Sign Dataset , an advanced dataset designed to support cutting-edge research and development in multimodal machine learning, speech recognition, and accessibility technologies. This dataset is uniquely tailored to the needs of DHH individuals, encompassing synchronized data across three key modalities:

**Audio:** High-resolution recordings of spoken Greek.

**Video:** Detailed lip movement recordings for speech alignment tasks.

**Text:** Precise transcriptions with word-level timestamps.

Additionally, the dataset integrates synchronized **sign language translations**, marking a significant milestone in the creation of AI tools for accessibility. This resource aims to support the development of technologies that improve accessibility in the tourism sector, such as real-time captioning systems, interactive translation applications, and educational tools for inclusive training.

By addressing the specific needs of DHH tourists and professionals, the **GLaM-Sign Dataset** serves as a foundation for pioneering advancements in accessibility, ensuring that the benefits of modern AI technologies extend to all individuals, regardless of their abilities.

## 2 Objectives and innovations of the dataset

The GLaM-Sign Dataset was developed with a clear objective: to advance accessibility technologies by addressing gaps in multimodal machine learning and speech recognition, particularly for the Greek language. Designed to empower the development of AI tools supporting Deaf and Hard-of-Hearing (DHH) individuals, the dataset provides synchronized data across audio, video, and textual modalities, complemented by integrated Greek Sign Language (GSL) translations. Its applications span automatic speech-to-sign translation systems, lip-reading AI, and real-time subtitle generation, offering significant advancements in accessibility within the tourism industry and beyond.

Globally, datasets such as GRID and TCD-TIMIT corpus [3], [4] have laid essential foundations for multimodal AI applications, predominantly in English-speaking contexts. For example, the GRID Corpus delivers approximately 34,000 utterances for controlled speech research, while TCD-TIMIT provides over 7,000 utterances tailored for audio-visual speech recognition tasks TIMIT. However, these datasets lack linguistic diversity, phonetic variation, and integration of sign language components, limiting their adaptability for non-English languages. The GLaM-Sign Dataset addresses these gaps by offering culturally and linguistically tailored data, along with innovative GSL integration. This approach enriches accessibility research while setting a benchmark for inclusive and representative datasets catering to underserved linguistic communities.

Significant research efforts have focused on accessibility for DHH individuals through advanced computational tools and technologies. Among the groundbreaking contributions to this field are studies that specifically target the Greek language and GSL. [5] introduced a rule-based machine translation (RBMT) system to generate high-quality GSL glossed corpora, facilitating the work of professional translators. Expanding on these contributions, combined RBMT with statistical machine translation (SMT) to produce parallel Greek-GSL corpora, achieving promising results in specific domains such as weather reporting. [6] further presented a prototype Greek text-to-GSL conversion system, incorporating GSL grammar into educational platforms. Additionally, [7] developed realistic 3D animations of GSL fingerspelling, advancing digital accessibility tools. Lastly, [8] emphasized the importance of robust linguistic resources in advancing Greek-to-GSL translations, further enhancing accessibility applications.

The GLaM-Sign Dataset distinguishes itself by addressing linguistic diversity, phonetic complexity, and multimodal accessibility. While [3], MS2SL [9] and TCD-TIMIT [4] provide valuable contributions, neither incorporates sign language components or captures the cultural and linguistic nuances of non-English languages. In contrast, the GLaM-Sign Dataset integrates synchronized audio, video, text, and GSL data, specifically designed to support AI tools for DHH individuals. This comprehensive multimodal framework, tailored to Greek linguistic structures and phonetic intricacies, establishes a robust foundation for research and innovation in accessibility technologies, particularly in underserved linguistic contexts.

Inspiration for this dataset also draws from pioneering projects in multimodal AI. Efforts like MS2SL, which focuses on continuous sign language production from multimodal inputs [9], and VALOR [10], which unifies vision, audio, and language for end-to-end understanding [10], align closely with accessibility goals. Advances in gloss-free sign language translation [11] highlight the value of reducing dependency on pre-existing gloss annotations, making translations more natural and efficient. Similarly, datasets such as WavCaps [12], focusing on audio-captioning, and frameworks like CALM [13], which align audio and language representations through contrastive learning [13], further emphasize the transformative potential of multimodal integration. Broader applications, such as robotic navigation systems leveraging multimodal data [14], showcase the versatility of such resources. By combining the unique aspects of the Greek language with innovations from these global efforts, GLaM-Sign Dataset positions itself as a trailblazing resource for AI-driven inclusivity, especially in underrepresented linguistic domains.

## 3 Description of the GLaM-Sign Dataset

The GLaM-Sign Dataset was developed as part of the FEELIT project , a groundbreaking initiative focused on enhancing accessibility and inclusivity within the tourism sector. Alongside the development of innovative educational materials—spanning three comprehensive chapters and enriched with visuals to improve comprehension—the project created this dataset to support research in accessibility technologies and multimodal AI development. The FEELIT project exemplifies the integration of theory, best practices, and cutting-edge technology to empower tourism professionals and Deaf and Hard-of-Hearing (DHH) individuals alike.

The FEELIT project has developed high-definition, video-based educational material specifically designed for Deaf and Hard-of-Hearing (DHH) learners. These videos enhance accessibility by providing sharper visuals, improved visibility of sign language interpreters, and reduced cognitive load. Spanning over 30 hours of content, the material serves to train both tourism professionals and DHH individuals. It is made accessible via a cloud-based platform, ensuring broad flexibility and ease of use.

Fig.1 illustrates the video layout, which features a signer alongside Greek subtitles positioned at the bottom. The subtitles are provided as external files in SRT format, allowing for easy customization by the end user.

This advanced GLaM-Sign Dataset integrates synchronized multimodal data, including high-resolution audio recordings, detailed video footage, and precise textual transcriptions, tailored for applications in speech recognition, lip-reading systems, and AI accessibility tools. It features 15,166 utterances totaling 279,042 words , with a vocabulary of 17,989 unique terms. The average utterance length is 18.41 words , covering a range from single-word phrases to multi-sentence expressions. These features make it suitable for applications requiring linguistic diversity and phonetic precision, aligning with global dataset standards for AI research.

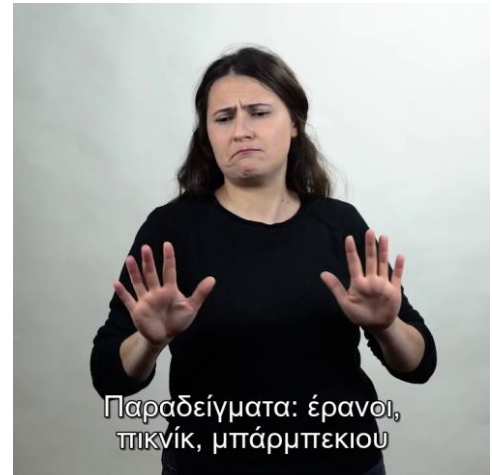

Fig.1: Sign Language Video Layout

While innovative, the dataset faces challenges such as minor misalignments in word-level timestamps and occasional phonetic errors. Despite this, evaluations show a 92% accuracy in automatic subtitle generation, with ongoing refinements needed to enhance precision and usability.

A high-resolution frame from a lip-reading dataset (Fig.2), showing a person speaking with clear facial visibility. The image focuses on the speaker's lips and facial expressions, ensuring optimal quality for visual speech recognition analysis. The frame is intended for training and evaluating models in the field of lip-reading and related accessibility technologies.

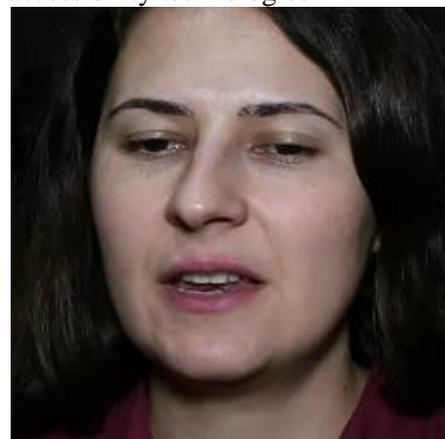

Fig.2: Frame from a lip-reading dataset highlighting clear facial and lip movement for visual speech recognition

Uniquely tailored to the Greek language, the GLaM-Sign Dataset bridges critical gaps in non-English linguistic resources. In contrast to globally recognized datasets like GRID (34,000 utterances) and TCD-TIMIT (7,000 utterances), this resource addresses the linguistic and phonetic complexities of Greek, offering a foundation for advancing regional and global accessibility technologies. It stands as a benchmark for inclusive AI research, fostering innovation in both linguistic and cultural contexts.

## 4 The Creation of the GLaM-Sign Dataset

The GLaM-Sign Dataset is a comprehensive resource designed to drive research and development in multimodal machine learning, speech recognition, and accessibility technologies. Focused on the Greek language, this dataset delivers high-quality, synchronized data across four key modalities:

**Speech:** High-resolution audio recordings of spoken Greek.

**Lip Movement:** Video recordings capturing detailed lip movements for tasks such as lip-sync and speech alignment.

**Subtitles:** Precise text transcriptions with timestamps, synchronized with the audio and video.

**Sign Language:** High-quality video translations in Greek Sign Language (GSL), providing an additional layer of accessibility for Deaf individuals.

For each video in the dataset, the spoken Greek content is complemented by synchronized Greek Sign Language translations, high-resolution subtitles, and corresponding audio narration. This multimodal approach ensures that the dataset serves as a versatile resource for applications in speech recognition, sign language translation, and accessibility technologies.

Developed as part of the FEELIT project, this dataset was created to support the development of AI-driven accessibility tools specifically tailored to the tourism sector. With over **30 hours of content**, it is designed to train both tourism industry professionals and Deaf or Hard-of-Hearing (DHH) individuals. These tools aim to enhance inclusivity by enabling DHH tourists and professionals to engage more effectively, fostering a more equitable and accessible travel experience.

By integrating sign language alongside spoken Greek and subtitles, the GLaM-Sign Dataset represents a significant leap in accessibility research. It bridges the gap between advanced AI capabilities and the unique linguistic and cultural needs of the Greek-speaking and DHH communities, setting a new standard for inclusive multimodal datasets.

## 5 Multimodal Dataset Preparation flow

**Code Workflow:** The preparation of the multimodal dataset involves a Python script (crop_video.py) (GitHub[1]) that processes video files to extract and crop facial regions using computer vision techniques. The process can be outlined in the following stages:

**Face Location Estimation:** The script utilizes the face_recognition library to estimate the average location of a face across all frames in a video. It samples frames at regular intervals and calculates the average bounding box of the face.

**Cropping and Resizing:** The video is revisited frame-by-frame. Each frame is cropped based on the calculated average face location, with additional padding determined by a user-defined ratio.

The cropped region is resized to 256x256 pixels while maintaining aspect ratio. The resized face is centered on a black background of the same resolution.

**Video Compilation:** The processed frames are saved into a new video file using the cv2.VideoWriter object, ensuring consistent quality and dimensions.

**Command-line Usage:**
--input: Path to the input video.
--output: Path to save the processed video.
--crop_ratio: Padding ratio around the face (default is 10%).
For example: python face_crop.py --input input_video.mp4 --output face_crop.mp4 --crop_ratio 0.1

Fig.3 illustrates the video layout, which On the left, a full-size video; on the right, the final cropped video focusing on the face and lips. The original frame shows the raw video frame before processing. The cropped frame displays the face region, centered and resized within a 256x256 frame.

**Audio Processing with FFmpeg:** Two FFmpeg commands were used during the dataset preparation:

**Extract Audio:** The audio stream was extracted from the original video into an *.m3a file:
"ffmpeg -i input_video.mp4 -vn -acodec copy extracted_audio.m3a"

**Integrate Audio:** The audio was re-integrated into the processed video:
"ffmpeg -i cropped_video.mp4-i extracted_audio.m3a -c:v copy -c:a aac -strict experimental final_video_with_audio.mp4"

---

[1] https://github.com/dkourem/Greek-Multimodal-Speech-Dataset-Corpus-v1

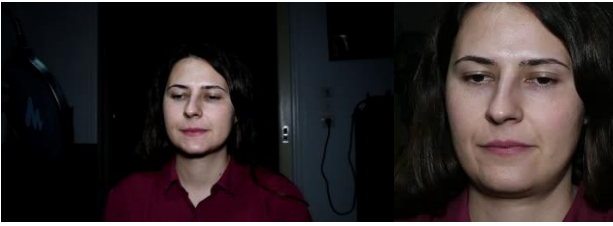

Fig.3: On the left, a full-size video with a resolution of 1280x720 pixels; on the right, the final cropped video focusing on the face and lips, resized to 256x256 pixels.

These images serve as visual evidence of the dataset's quality and the accuracy of face-centric cropping.

## 5.2 Subtitle Generation Using Whisper

Code Workflow The process of generating subtitles involves a Python script
(whisper_transcribe_per_sent_and_per_word.py)
that uses the Whisper model for audio transcription. This workflow highlights the critical steps and considerations:

**Audio Transcription:** The Whisper model (large-v2) transcribes audio files into text, with support for word-level timestamps.

The transcription includes detailed segmentations, each containing start and end times.

**Saving Transcription Data:** The output is saved in a JSON format, including both sentence-level and word-level timestamps.

Note: Word-level timestamps may not always be precise and require manual review for accuracy.

**Generating Subtitle Files:** Sentence-level subtitles: Timestamps and text are formatted into .srt files, representing complete sentences. Word-level subtitles: Each word is timestamped and saved in .srt format for finer synchronization.

Command-line Usage The script supports the following arguments:
--input: Path to the input audio or video file.
--output: Directory to save the JSON and subtitle files.
Example command:
Python whisper_transcribe_per_sent_and_per_word_v3.py --input audio_file.mp4 --output .subtitles

**Output Formats:**
**JSON File:** Contains the transcription result, including timestamps for words and sentences.
**Sentence-level SRT:** Synchronizes complete sentences with their timestamps.
**Word-level SRT:** Synchronizes individual words with their timestamps.
Manual Review Required: The word-level timestamps, while detailed, often require manual adjustments. Manual corrections were performed using the free software Subtitle Edit to ensure accuracy and synchronization[2] (Fig.4).

Phonetic alignment models are crucial for accurately mapping spoken words to their corresponding text or visual representations, enabling accessibility tools such as real-time captioning, lip-reading systems, and speech-to-sign translation to function with high precision, particularly in scenarios involving diverse linguistic and phonetic variations.

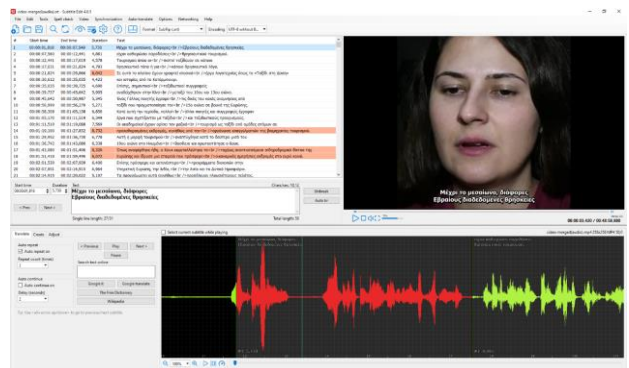

Fig.4: Manual adjustments were necessary for the word-level timestamps to ensure synchronization and accuracy. Corrections were carried out using the free software Subtitle Edit

## 5.2 Handling Phonetic Errors Using N-gram Models

While Whisper achieves high transcription accuracy, certain phonetic errors can occur. For example, the correct word "ευρέως" might be transcribed as "εβραίως" due to phonetic similarity. To streamline corrections, we developed an N-gram model based on the original text corpus narrated by the actor. The process involves:

**Text Cleaning:**
Texts were preprocessed using a cleaning script (TextCleaningNormalization.py).
This step removes non-Greek characters, normalizes punctuation, and ensures clean input for training.

**N-gram Model Training:**
The script NgramModelCreation.py generates an n-gram model (bi-grams and tri-grams) from cleaned text using the Python platform. Specifically, it leverages key Python libraries such as 'nltk' for natural language processing (tokenization and n-gram generation), 're' for text cleaning using regular expressions, and 'collections' for efficient counting and grouping of n-grams. The cleaned text is loaded, converted into a list of words, and used to extract the desired n-grams. The output includes n-gram frequencies, which can be utilized for linguistic pattern analysis or predictive model development. The N-gram model is saved as a serialized object for future use.

---

[2] https://github.com/SubtitleEdit/subtitleedit/releases

**Phonetic Similarity Analysis:**

A function (PhoneticSimilarityFunction.py) calculates phonetic distances based on the International Phonetic Alphabet (IPA) representation.

Each word is converted into its phonetic form using a phonetics library, such as Metaphone, which maps similar sounding words to the same or similar phonetic codes. For example, the IPA representation of "ευρέως" is /eˈvre.os/, while "εβραίως" is represented as /eˈvre.i.os/. These phonetic forms help identify subtle differences and similarities for accurate error detection. The weighted phonetic distance between the transcribed word and possible replacements is computed. This includes:

Length Difference: A weight is applied to penalize large discrepancies in length between phonetic representations.

Character Comparison: Each character of the phonetic codes is compared; mismatches are penalized with a fixed score.

**Error Detection and Correction:**

Subtitles are analyzed using an error-checking script (SubtitleChecking-Phonetic.py).

For example, if "εβραίως" is detected in a tri-gram context that frequently includes "ευρέως," the model suggests corrections.

Phonetic similarity analysis confirms "ευρέως" as the most likely replacement.

Example Correction Workflow:

Flagging: The word "εβραίως" is flagged during subtitle analysis.

Tri-gram Matching: Surrounding words are matched against tri-grams in the N-gram model, with "ευρέως" emerging as a high-probability candidate.

Phonetic Verification: The IPA-based phonetic similarity analysis validates "ευρέως" as the most contextually and phonetically suitable replacement.

For example, the IPA representations of "ευρέως" (/eˈvre.os/) and "εβραίως" (/eˈvre.i.os/) are compared. The phonetic similarity function calculates a weighted distance by:

Identifying differences in length, where the additional phoneme /i/ in "εβραίως" increases the score. The phonetic distance is calculated using the following equation:

$$D = w_1 \times |L_1 - L_2| + w_2 \times \Sigma \, \delta(c_{1i}, c_{2i})$$

Where:

$L_1$ and $L_2$ are the lengths of the phonetic representations.

$w_1$ is the weight for length difference.

$\delta(c_{1i}, c_{2i})$ is a mismatch penalty for characters $c_{1i}$ and $c_{2i}$.

$w_2$ is the weight for character mismatches.

For example, comparing "ευρέως" (/eˈvre.os/) with "εβραίως" (/eˈvre.i.os/):

$|L_1 - L_2| = 1$ due to the additional phoneme /i/ in "εβραίως."

A character-by-character comparison finds one mismatch (absence of /i/ in "ευρέως").

The calculated distance determines that "ευρέως" and "εβραίως" have close phonetic properties, making "ευρέως" a likely correction.

This approach ensures that "ευρέως" is recognized as the most phonetically and contextually accurate replacement.

Updating Subtitles: The flagged word is replaced with "ευρέως" in the subtitle file.

This pipeline significantly reduces manual correction time while maintaining high transcription accuracy by leveraging both context and phonetic properties.

# 6 Accuracy of Word-Level Timestamps and Future Work

The GLaM-Sign Dataset has demonstrated strong capabilities in sentence-level timestamping, particularly in cases with clear pauses or well-defined breaks in audio. However, at the word level, the accuracy decreases, with correct timestamps achieved for approximately 30–40% of words in preliminary analyses. This issue stems from challenges such as overlapping speech, rapid dialogue, and unclear enunciation.

To address these challenges, future iterations of the dataset will incorporate advanced alignment techniques and additional annotated datasets. These enhancements will improve the precision of word-level timestamps and provide a more robust framework for model training. Planned refinements include retraining the Whisper model and developing complementary synchronization models that leverage reinforcement learning and phonetic alignment algorithms . These improvements aim to optimize temporal accuracy and usability, ensuring that the dataset remains a cutting-edge resource for multimodal AI applications.

The dataset has already undergone rigorous spelling and typographical corrections to ensure readiness for initial applications. However, refining word-level timestamps remains an ongoing effort, necessitating meticulous review and targeted corrections to address phonetic ambiguities and enhance synchronization. The use of iterative feedback within neural networks, as demonstrated by prior research [15], will also be explored to refine alignment across audio, video, and text modalities.

To address the challenges of word-level timestamp accuracy, we envision a collaborative approach involving both technical experts and accessibility advocates. By engaging researchers in linguistics, computer vision, and accessibility technology, the dataset can benefit from novel alignment techniques and real-world feedback. These collaborations will enable iterative refinements, ensuring that the dataset evolves to meet diverse research and application needs.

## 7 Conclusion

The creation of the GLaM-Sign Dataset marks a significant milestone in accessibility research and multimodal AI development. Far more than a collection of data, this dataset serves as a sophisticated and versatile resource, integrating synchronized high-resolution audio, detailed video recordings, precise textual transcriptions, and Greek Sign Language (GSL) translations. By addressing the unique needs of Deaf and Hard-of-Hearing (DHH) individuals, it fills a critical gap in accessibility-focused AI datasets, providing unparalleled multimodal alignment and cultural specificity.

The dataset enables groundbreaking applications, including real-time sign language translation, speech-to-sign systems , and enhanced subtitle synchronization . These technologies empower DHH individuals and foster inclusivity in tourism by addressing key challenges such as inaccessible voice-only announcements, insufficiently trained staff, and limited accessible services. For instance, tools developed from this dataset can facilitate interactive translation apps , visual itinerary planners , and AI-driven virtual concierges , breaking communication barriers and enriching the travel experience for DHH users.

While the dataset already represents a robust foundation, ongoing work is focused on addressing current limitations, such as word-level timestamp precision and linguistic constraints. Future iterations aim to implement a hybrid framework that combines machine learning-based phonetic alignment models with human-in-the-loop correction systems, leveraging automated systems for initial alignment and human expertise to refine edge cases. Incorporating advanced tools like reinforcement learning and context-aware neural networks will enhance word-level synchronization and usability, creating a resource tailored for both research and real-world applications.

Additionally, the dataset's framework is designed with scalability in mind, making it adaptable for other languages and cultural contexts. By inviting researchers and developers to adapt the methodology, the dataset has the potential to foster global accessibility solutions. Partnerships with academic institutions and accessibility-focused organizations will amplify its impact, ensuring its relevance across diverse linguistic and cultural challenges.

To encourage collaboration and innovation, the dataset's preparation workflows and scripts will be openly available. This transparency will invite contributions from the wider research community, including students, who can explore and expand upon approaches to alignment, timestamp precision, and multimodal integration. These efforts will ensure the dataset remains a living, evolving resource.

In conclusion, GLaM-Sign Dataset exemplifies the transformative potential of multimodal AI in accessibility. It not only lays the groundwork for innovations in tourism but also offers a template for developing tools that empower underrepresented communities across sectors such as education, healthcare, and public services. By fostering inclusivity, transparency, and collaboration, this work sets a benchmark for ethical and empathetic AI development, paving the way for a universally accessible future.

---

[i] https://doi.org/10.5281/zenodo.14610495